# An Effective Semi-supervised Divisive Clustering Algorithm

**Teng Qiu, Yongjie Li***

Key Laboratory of NeuroInformation, Ministry of Education of China, School of Life Science and Technology, University of Electronic Science and Technology of China, Chengdu, 610054, China
*Corresponding author. Email: liyj@uestc.edu.cn

**Abstract:** Nowadays, data are generated massively and rapidly from scientific fields such as bioinformatics, neuroscience and astronomy to business and engineering fields. Cluster analysis, as one of the major data analysis tools, is therefore more significant than ever. Here, we propose an effective **S**emi-supervised **D**ivisive **C**lustering algorithm (**SDC**). Data points are first organized by a minimal spanning tree. Next, this tree structure is transitioned to the in-tree structure, and then divided into sub-trees under the supervision of the labeled data. In the end, all points in the sub-trees are directly associated with specific cluster centers. SDC is fully automatic, non-iterative, involving no free parameter, insensitive to noise, able to detect irregularly shaped cluster structures, applicable to the data sets of high dimensionality and different attributes. The power of SDC is demonstrated on several datasets.

## 1 Introduction

Cluster analysis is old and fundamental in computation area, aiming at classifying data points into categories based on their similarities. Diverse experimental data ranging from microarray gene expression data in biology to spectrum data in astronomy require to be clustered to signal meaningful correlation of the data. Massive documents or images on internet are also needed to be effectively organized so as to promote the efficiency of search engines.

Clustering method as K-means (*1*) is popular for its simplicity, yet sensitive to noise and initialization and thus is limited by the lack of reliability. Hierarchical clustering (HC) (*2*) is simple and intuitive and thus widely used especially in biology (*3*), whereas it needs a large computation (*4*) and its result is variable to a set of similarity measures between clusters. Moreover, the cluster number for the above methods needs to be prespecified (e.g., K-means) or determined by a threshold (e.g., HC). Some other well-known algorithms either involve complex optimization and postprocessing (*5*), or have limited range of applications such as the distribution (*6*) or the attribute of data (*7, 8*). Although affinity propagation (AP) (*9*) has much better performance than K-means and the cluster number is determined automatically, it is not good at detecting nonspherical clusters (*10*).

Recently, two effective clustering algorithms (*10, 11*) were proposed, which can together form a pool of clustering methods based on the in-tree structure (*11*). But they involve a free parameter.

## 2 Principle and features of SDC

All the above problems are not involved in the proposed clustering algorithm

SDC. Provided with several labeled data points, SDC can find the underlying clustering structure in four steps (see flowchart in Fig. 1A , example in Fig. 1 D to H):

First, SDC connects all data points (or nodes) by the minimal spanning tree (**MST**) (*12*, *13*) (Fig. 1E), for which, the sum of the edge lengths (i.e., the distances of the connected nodes) is minimum among all possible connected graphs.

Then, SDC selects any node as the ancestor node (or root node) R, and successively identifies its offspring: the children nodes (directly connected with R), children's children, etc. until all nodes are considered. Each parent node may have more than one child node, whereas, when tracing backward, each child node has only one parent, one grandparent, etc. and the common ancestor R in the end. After determining the "Family Relationship" of each pair of connected data points on MST, each child node is treated as an initial node that points to its parent node, i.e., each undirected edge on MST becomes directed (illustrated in Fig. 1B). Consequently, the MST (Fig. 1E) becomes an directed tree (Fig. 1F), or more precisely, the in-tree (**IT**) (*14*), for which, along the edge direction, each node has one and only one path to reach the root node.

In the third step, the IT structure is divided into several sub-trees by removing some undesired edges based on two simple divisive rules (illustrated in Fig. 1C): (i) each sub-tree must contain at least one labeled node; (ii) impure sub-trees (i.e., the sub-trees containing different labeled nodes) should divide. Under these rules, the edges are explored in decreasing order of their lengths (the undesired edges are in general much longer). In Fig. 1F, for the five longest edges, only the 1st and 5th longest edges are removed. Consequently, three sub-trees are obtained (Fig. 1G), each of which is still an IT structure with one root node (the initial node of the removed directed edge). Then, the exploration stops.

Finally, searching along the direction of edges, all data points will converge at the root nodes of different sub-trees (Fig. 1H). The nodes connected with the same root nodes are assigned in the same clusters. If there are clusters containing the same labeled nodes, they should be of the same category, thus, the cluster number is sure to be the category number of the labeled nodes.

The whole process is fully automatic. No free parameter is involved. The cluster number is not needed to specify in advance. No such application constraints as shape, attribute and dimensionality are imposed on the test data. SDC is also to some degree insensitive to noise, especially to outliers.

Although SDC requires some of the raw data points to be previously labeled, one advantage is accompanied as the participation of the labeled data, i.e., the clustering result could be more reliable than that of the unsupervised clustering methods, since the result should at least be consistent with the category of the labeled data.

## 3 Experiments

Figure 2A provides a synthetic data set, with several clusters differing in size, shape and density, and contaminated by noise. Nine out of 5404 points are labeled in different categories (denoted by different colors). It is easy for our eyes to spot all these clusters, whereas previously hard for the computer by virtue of any clustering algorithm (*15*). Now, with the help of our algorithm, computer can be as intelligent as us (Fig. 2B).

Previously, it should be rather tedious for scientists to label 8124 mushrooms as either poisonous or edible (Fig. 3A) (*16*) one by one. However, scientists could just label only a small number, and for the remaining, a satisfactory prediction can be made by our algorithm (Fig. 3B), which is actually close to the work of experts. The more labeled data there are, the more reliable the prediction is.

Thanks to the in-tree structure obtained in step 2, each exploration of the edge in step 3, though involving a judgment of whether to cut it or not, is not time-consuming. Suppose one edge is removed, all nodes can first be associated with root nodes, a process same as step 4, then it is easy to judge, according to the divisive rules, whether there is one labeled node or whether the labeled nodes are the same among the offspring nodes of each root node. Since the process of finding root nodes in step 4 can be extremely fast (*11*), the time cost by step 3 is also negligible (Fig. 3C).

We also applied SDC to cluster the Olivetti Face Dataset (*17*). Unlike mushroom dataset, this dataset has much less instances (Fig. 4A, 400 grayscale 112 pixel by 92 pixel face images) than features (each face is treated as a long vector of 10304 features). Since there are only 400 data points in the 10304-dimensional space, the distribution of the data points can be extremely sparse, and thus the task of clustering them is quite challenging (*10*). However, it's not so troublesome to our algorithm, provided that several images are labeled, identical if they are from same subjects, and different otherwise. Figure 4B shows the case of two labeled images from each subject. Consequently, the remaining images can be clustered exactly into 40 clusters with only 15 wrong classifications (Fig. 4C), which is better than the result recently reported in (*10*), though with additional need for the labeled data. Moreover, like the case of clustering mushrooms, more labeled data can result in more reliable clustering result (Fig. 4D), since more labeled data can lead to a more elaborate division.

## 4 Discussions

SDC inherits the advantages of both the MST, IT structures and the semi-supervised learning strategy. The MST structure has long before attracted people's attention, due to its fascinating characteristics (*18*). For example, the "minimal principle" of the MST provides an effective and universal way to organize data points into a graph structure regardless of the distribution and attribute of the data points, with no need to set any parameter. This "minimal principle" is also in close conformity with the "proximity principle" of "Gestalt" perceptual organization (*19*) and the "sparse coding" (*20*) feature of our nervous system. The IT structure has great beauty in its order (directed, cycle-free), certainty (every node has one and only one path to reach the root node) and efficiency (its evolution can be parallel and thus fast). The power of the IT structure has been demonstrated in one recently proposed clustering algorithm (*11*). Moreover, the semi-supervised learning (*21*), a combined use of both the labeled and unlabeled data, has attracted increasing interest and shown its superiority over the supervised and unsupervised learning which only relies on either labeled data (supervised) or unlabeled data (unsupervised). For our SDC, a natural transition from the MST to the IT structure and an effective semi-supervised cutting mechanism (i.e., the divisive rules) make it simple, fast, effective and reliable.

SDC presents such an interesting learning behavior of computer: the raw materials (e.g., mushrooms and face images) are first organized in a sparse form, i.e., the MST structure (step 1), and then effectively transitioned to a more sparse one, the IT structure (step 2). Until the computer is informed that some of the raw materials are different (example-based learning), the above IT structure starts to evolve or divide, so as to be in line with the known (step 3). The more examples computer learns from, the more reliable this evolution can be. Consequently, the unknown materials are explicitly associated with the known, in the light of which, all the unknown are also lighted up. In other words, the computer becomes more knowledgeable.

Since the proposed algorithm, SDC, provides an effective way to indirectly derive the IT structure from the MST structure, it can be viewed as a new (the 6th) member of the IT clustering family proposed in (*11*).

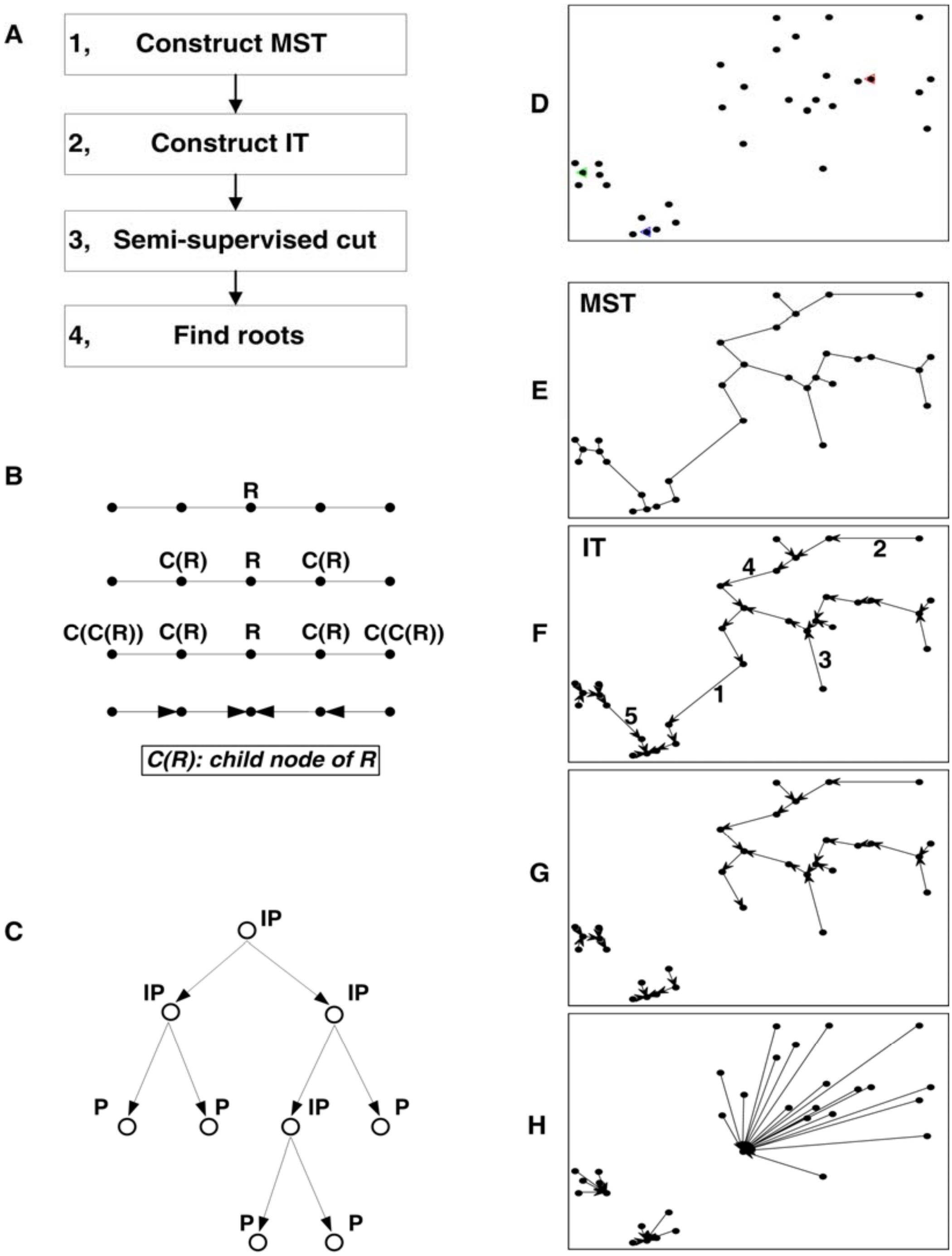

**Fig. 1. How the proposed algorithm SDC works.** (**A**) The flowchart of SDC. (**B**) An illustration for step 2. Row 1: a MST structure is constructed for five nodes and one of them (middle) is selected as the root node R. Row 2 and 3: the children nodes and children' children nodes of node R are successively identified. Row 4: each child node points to its parent node. (**C**) An illustration for the divisive rules in step 3, where the sub-tree (denoted by circle) containing the same labeled nodes is called pure (P) one, otherwise impure (IP) one. (**D**) One synthetic dataset. The labeled data are denoted by three triangles, different colors of which represent different categories of the labeled data. (**E** to **H**) respectively corresponds to step 1 to 4 in (A).

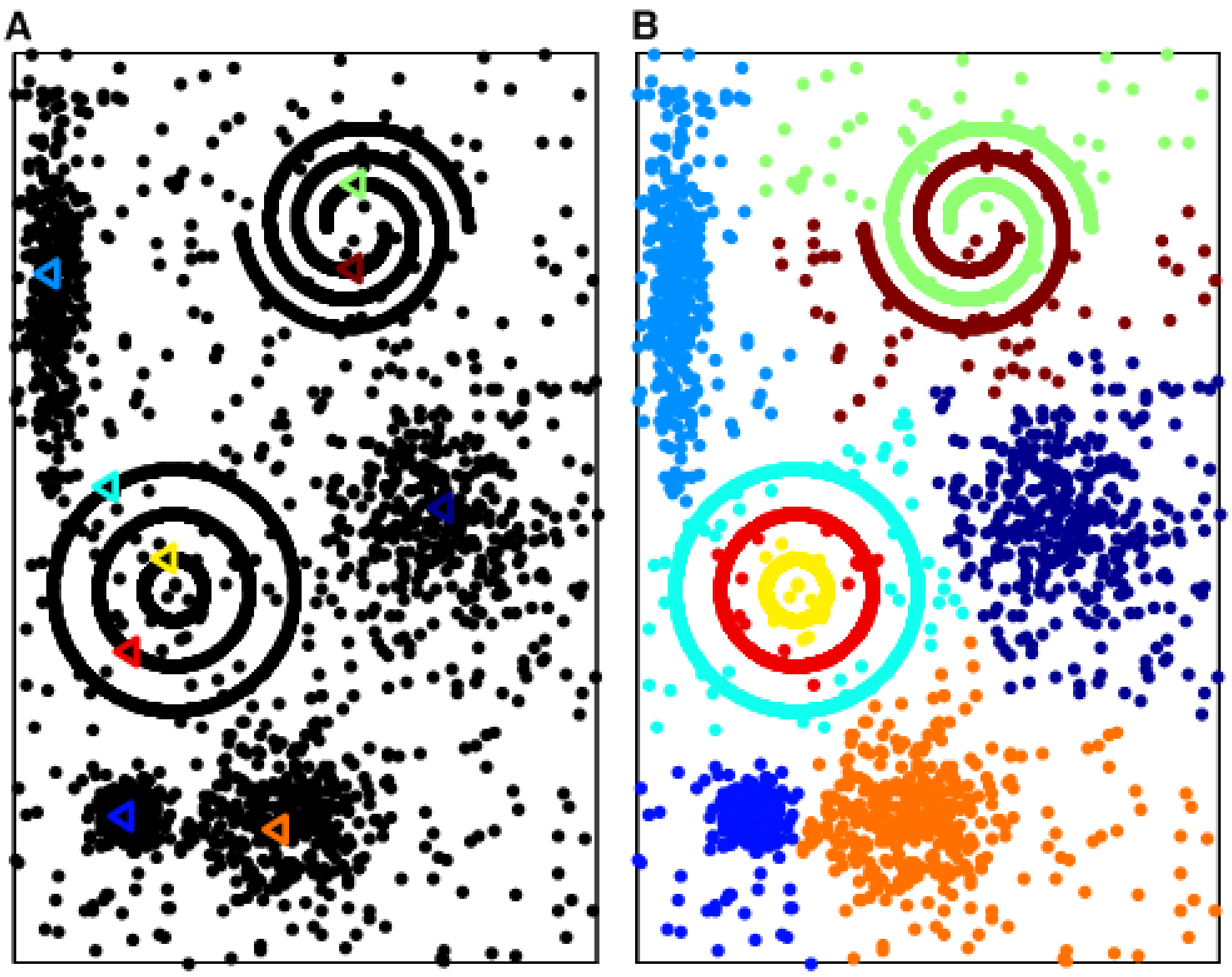

**Fig. 2. Clustering synthetic data points.** (**A**) The synthetic dataset. Triangles denote the place where the nine labeled data points locate. Different colors of these triangles represent different categories of the labeled data. (**B**) Clustering result. Points in the same colors belong to the same clusters.

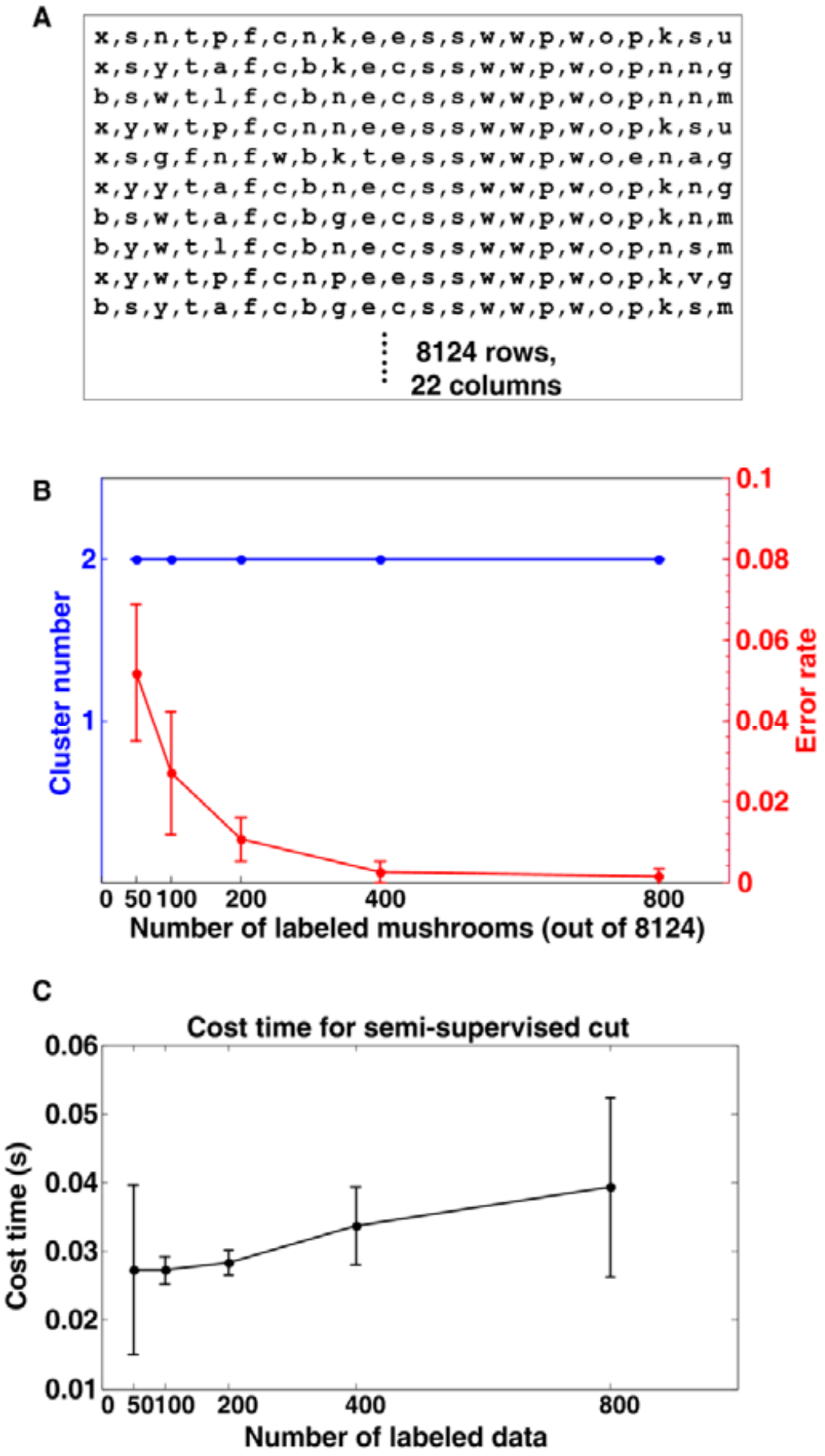

**Fig. 3. Clustering mushrooms.** (**A**) A small portion of the mushroom data set. Each mushroom (i.e., each row) is featured by 22 characters. The distance between each pair of mushrooms is measured by the number of positions (or columns) having different elements. (**B**) A plot of the cluster number (blue) and error rate (red) versus the number of labeled mushrooms, respectively. Error rate (same in Fig. 4) is the ratio of the number of the falsely assigned instances to that of the unlabeled instances. (**C**) A plot of the time cost for the semi-supervised cutting versus the number of labeled data. In (B and C), the points indicate the means and the error bars indicate the standard errors over 20 random tests.

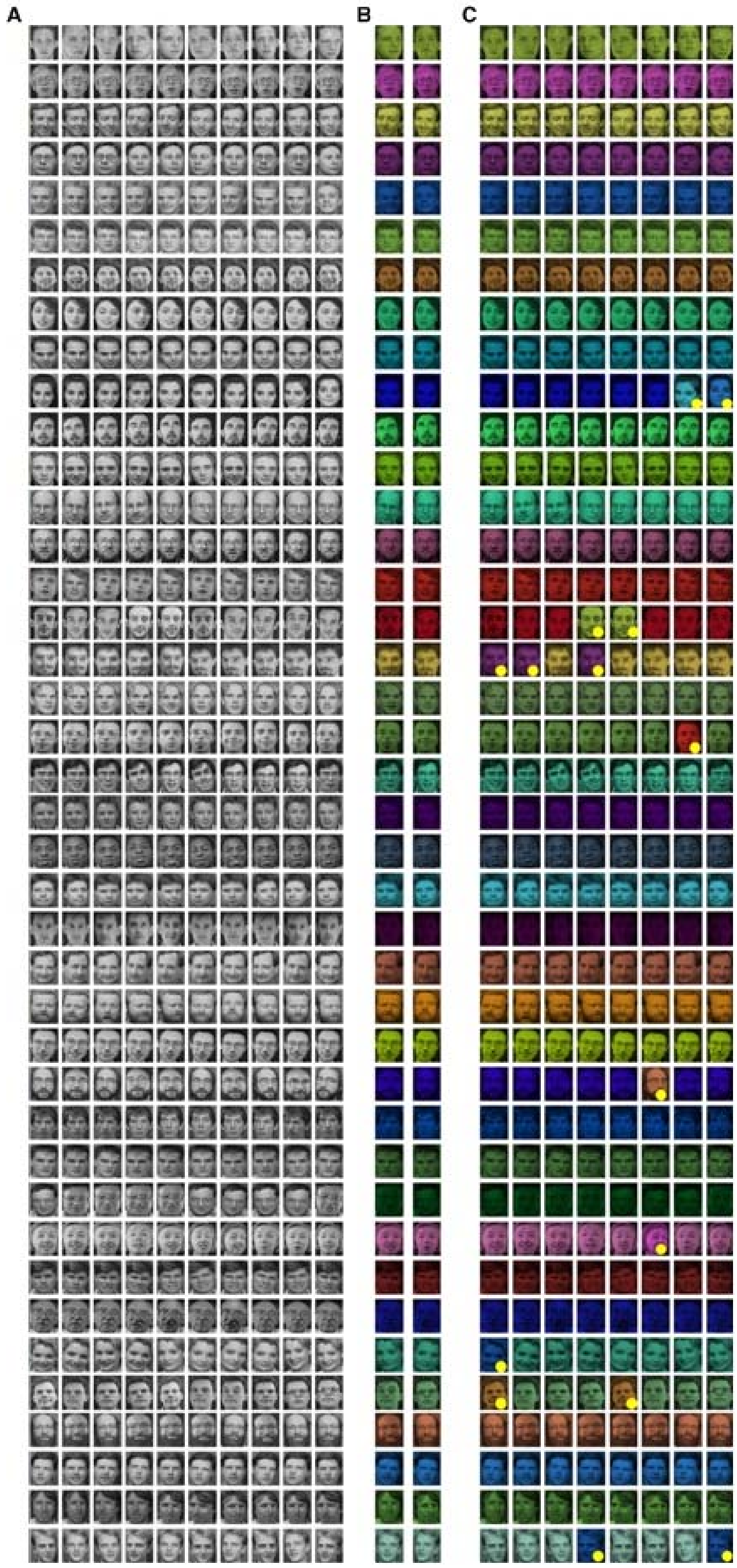

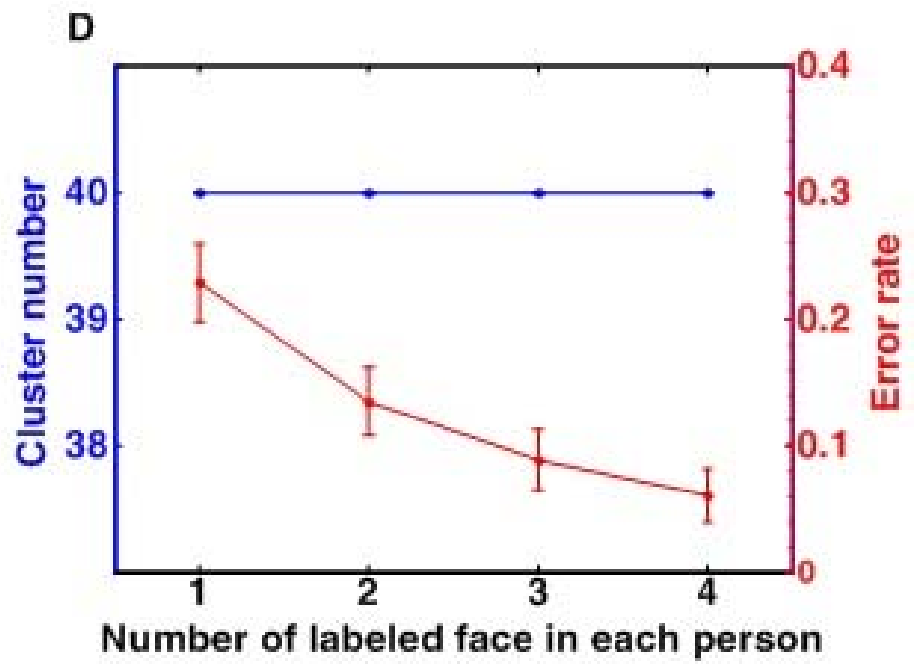

**Fig. 4. Clustering faces.** (**A**) Four hundred grayscale faces of 40 subjects. Each row shows 10 varying images of one subject. Each image is treated as a 10304-dimensional vector. The distance between each pair of faces is measured by the Euclidean distance. (**B**) The labeled faces. Two faces are randomly selected from each subject. (**C**) Clustering assignments (denoted by different colors) of the unlabeled images. Only fifteen images (denoted by the yellow dots) are falsely assigned. (**D**) A plot of the cluster number (blue) and error rate (red) versus the number of labeled faces in each person, respectively. The points indicate the means and the error bars indicate the standard errors over 200 random tests.